\documentclass[journal]{IEEEtran}
\IEEEoverridecommandlockouts
\usepackage{cite}
\usepackage{amsmath,amssymb,amsthm,amsfonts}
\usepackage{algorithmic}
\usepackage{epstopdf}
\usepackage{textcomp}
\usepackage{xcolor}
\usepackage{url}
\usepackage{algorithmic, algorithm}
\newcommand{\st}{\mathrm{s.t.}}

\newcommand{\tr}{\mathrm{tr}}

\usepackage{longtable,multirow,colortbl,booktabs}

\usepackage{graphicx,subfig}
\usepackage{float}

\usepackage{graphicx}
\usepackage{subfig}

\ifCLASSINFOpdf
\else
\fi
\begin{document}
\title{Fast Asymmetric Factorization for Large Scale Multiple Kernel Clustering \\
	
	\thanks{This work is supported by the National Natural Science Foundation of China grants, 62376146, 62176001. (Corresponding author: Lei Duan)}
}

\author{Yan~Chen, Liang~Du,~\IEEEmembership{Member,~IEEE},
        Lei~Duan,~\IEEEmembership{Member,~IEEE,}% <-this % stops a space
\thanks{Yan Chen and Lei Duan are with the College of Computer Science, Sichuan University, Chengdu 610065, China (e-mail: chenyan557@stu.scu.edu.cn; leiduan@scu.edu.cn).}
\thanks{Liang Du are with Institute of Big Data Science and Industry, Shanxi University, Taiyuan 030006, China (e-mail: duliang@sxu.edu.cn)}}


\markboth{Journal of \LaTeX\ Class Files,~Vol.~14, No.~8, August~2015}%
{Shell \MakeLowercase{\textit{et al.}}: Bare Demo of IEEEtran.cls for IEEE Journals}


\maketitle

\IEEEpeerreviewmaketitle

\begin{abstract}
	Kernel methods are extensively employed for nonlinear data clustering, yet their effectiveness heavily relies on selecting suitable kernels and associated parameters, posing challenges in advance determination. In response, Multiple Kernel Clustering (MKC) has emerged as a solution, allowing the fusion of information from multiple base kernels for clustering. However, both early fusion and late fusion methods for large-scale MKC encounter challenges in memory and time constraints, necessitating simultaneous optimization of both aspects. To address this issue, we propose Efficient Multiple Kernel Concept Factorization (EMKCF), which constructs a new sparse kernel matrix inspired by local regression to achieve memory efficiency. EMKCF learns consensus and individual representations by extending orthogonal concept factorization to handle multiple kernels for time efficiency. Experimental results demonstrate the efficiency and effectiveness of EMKCF on benchmark datasets compared to state-of-the-art methods. The proposed method offers a straightforward, scalable, and effective solution for large-scale MKC tasks. The code for our method is publicly available at \url{https://github.com/YanChenSCU/EMKCF.git}.
\end{abstract}


\section{Introduction}
Kernel methods have been extensively studied to address nonlinear data clustering. However, traditional kernel methods often require selecting an appropriate kernel and its associated parameters, posing challenges in advance determination. To alleviate this, MKC has garnered significant attention, enabling the fusion of information from multiple base kernels for clustering. Existing MKC algorithms can be broadly categorized into two groups based on their fusion strategies: early fusion methods and late fusion methods.

The first category integrates multiple candidate kernels to estimate a consensus kernel matrix, aiming to capture the consensus cluster structure. Typical methods include MKKMMR \cite{mkkmmr}, Simple SMKKM \cite{SMKKM_TPAMI_2023}, and Localized SMKKM \cite{LSMKKM_2021}, with cubic complexity. Besides, Wang et al \cite{DPMKKM_tip_2022} propose DPMKKM to learn the discrete partition matrix directly, reducing time complexity to quadratic. Moreover, EMKC \cite{EMKC} and FAMKKM \cite{FAMKKM} generate two partitions for each kernel and integrate them for consensus learning, with linear time complexity. In recent years, several kernel graph methods have been introduced to enhance separability. These methods focus on learning an affinity graph from a kernel. However, the global kernel graph is memory-intensive for dense storage and computationally inefficient, while the local kernel graph, although sparse, is also inefficient for computing. However, these methods demand significant memory to store original dense kernels. For example, on the EMNIST dataset, a single dense kernel of 280,000 samples requires 584 GB memory, limiting their suitability for large-scale MKC clustering. 

The late fusion methods, exemplified by techniques like OPLFMVC \cite{liu2021one}, MKCCSA \cite{CSAMKC}, and LFPGR \cite{lfpgr_tnnls_2023}, initially extract multiple partition matrices from kernels and then utilize consensus learning to derive the final clustering solution, typically employing a linear fusion process. However, the generation of candidate partitions inevitably leads to the loss of certain information and is time-consuming when derived from individual squared kernel matrices. Moreover, relying solely on a single partition fails to fully exploit the complementary information offered by other kernels. Consequently, the performance of late fusion methods appears to be inferior in comparison to the early fusion counterparts.

The challenges encountered in large-scale Multiple Kernel Clustering (MKC) can be summarized into three primary aspects. Firstly, significant memory requirements arise due to the substantial storage demands of original dense kernel matrices, even for just a single kernel matrix. Secondly, the high time complexity involved in processing squared kernel matrices detrimentally impacts computational efficiency. Lastly, achieving efficiency in both memory and time costs should be addressed simultaneously rather than independently.

To address the simultaneous challenges of memory and time constraints, we propose a novel approach called Efficient Multiple Kernel Concept Factorization (EMKCF) tailored for large-scale MKC. Instead of relying on original dense kernel matrices, we introduce the construction of a new sparse kernel matrix inspired by local regression. Notably, computation of this sparse kernel can be performed on a single row within a single kernel, allowing independent construction by loading only one row into memory, rather than the entire dense kernel matrix. This approach optimizes both computation and storage for large-scale MKC. Furthermore, we extend orthogonal concept factorization to its multiple kernel counterparts to enhance efficiency. This extension aims to achieve consensus representation across all sparse kernels while generating kernel-dependent representations for each kernel. The factorization of both consensus and individual representations is embedded within a new framework. Additionally, we devise an alternative update algorithm with linear time complexity to update all variables. Experimental results on large-scale datasets showcase the efficiency and effectiveness of EMKCF compared to state-of-the-art techniques.

\section{The Proposed Method}
In MKC a dataset $\mathbf{X} = [\mathbf{x}_1, \mathbf{x}_2, \ldots, \mathbf{x}_n]$ in $\mathbb{R}^{d\times n }$ undergoes transformation via a mapping function $\phi : \mathcal{X} \mapsto \mathcal{F} $, enhancing separability in a higher-dimensional space $\mathcal{F}$ determined by the inner product $\left\langle\cdot, \cdot\right\rangle_{\mathcal{F}}$ governed by a kernel function $\mathcal{K}(\cdot, \cdot)$. Utilizing a set of candidate kernel functions $\{\mathcal{K}_r^o(\cdot, \cdot)\}_{r=1}^{m}$, we construct positive semi-definite kernel Gram matrices $\{\mathbf{K}_r^o\}_{r=1}^{m}$ ($\mathbf{K}_r^o \in \mathbb{R}^{n \times n}$) capturing pairwise similarities, essential for subsequent clustering analysis.
\subsection{Sparse Kernel Extraction}

Given an original kernel matrix $\mathbf{K}^o$, we identify the local clique of a sample $\mathbf{x}_i$ by selecting its top-$k$ nearest neighbors $\mathcal{N}_i =\{\mathbf{x}_{j_1},\mathbf{x}_{j_2},\ldots,\mathbf{x}_{j_k}\}$ for simplicity. The Nadaraya-Watson kernel estimator, a well-known non-parametric regression technique for nonlinear regression \cite{benedetti1977nonparametric}, computes the target label $y_i$ using the neighbors $\mathcal{N}_i$ of sample $\mathbf{x}_i$ according to the formula: $y_i = (\sum_{\mathbf{x}_{j} \in \mathcal{N}_i} \kappa_{ij}y_j)/(\sum_{\mathbf{x}_{j'} \in \mathcal{N}_i} \kappa_{ij'})$. Consequently, a local affinity matrix $\mathbf{S}$ is populated using the Localized Nadaraya-Watson Kernel Regression (LNWKR) coefficient as follows:
\begin{align} \label{lnwkr}
	s_{ij} = \left\{ \begin{array}{ll}
		\frac{\kappa_{ij}}{\sum_{\mathbf{x}_{j'} \in \mathcal{N}_i} \kappa_{ij'}}, &  \textrm{if } \mathbf{x}_j \in \mathcal{N}_i, \\
		0, &  \textrm{if } \mathbf{x}_{j} \notin \mathcal{N}_i.
	\end{array}
	\right.
\end{align}
It can be verified that $\mathbf{S} \geq 0, \mathbf{S} \mathbf{1}= \mathbf{1}$.
Given an asymmetric affinity graph \( \mathbf{S} \) from the LNWKR model, we first introduce the corresponding symmetric matrix \( \mathbf{A} = (\mathbf{S} + \mathbf{S}^{T})/2 \), and the diagonal degree matrix \( \mathbf{D} \) where its diagonal element \( \mathbf{D}(i,i) = \sum_{j=1}^{n}a_{ij} \), then we define the new sparse and neighbor kernel as follows,
\begin{align}\label{new_kerenl}
	\mathbf{K} &= (\mathbf{I} + \mathbf{D})^{-\frac{1}{2}} (\mathbf{I} + \mathbf{A}) (\mathbf{I} + \mathbf{D})^{-\frac{1}{2}}.	
\end{align}
It can be verified that the constructed matrix $\mathbf{K}$ is also a kernel matrix with its eigenvalues in \( [0, 1] \). As a consequence, we can build $m$ kernels $\{\mathbf{K}_r\}_{r=1}^{m} $ from the original kernels $ \{\mathbf{K}_r^o\}_{r=1}^{m}$.

%\subsubsection{Sparse Kernel Enhance Enhancement}

\subsection{Multiple Kernel Orthogonal Concept Factorization}
Given an arbitrary kernel matrix $\mathbf{K}$, the optimization problem known as KCF \cite{xu2004document}, described in Eq.~\eqref{kcf}, aims to minimize the squared loss while subject to non-negativity constraints on the low-dimensional factor matrices $\mathbf{U}$ and $\mathbf{H}$:
\begin{align}\label{kcf}
	\min_{\mathbf{U}, \mathbf{H}} \quad & \tr(\mathbf{K}) - 2 \tr(\mathbf{U}^T\mathbf{K} \mathbf{H}) + \tr(\mathbf{H}\mathbf{U}^T\mathbf{K}\mathbf{U}\mathbf{H}^T )  \\
	\text{s.t.} \quad & \mathbf{U} \geq 0, \quad \mathbf{H} \geq 0,\nonumber
\end{align}
It is noteworthy that the solution to Eq.~\eqref{kcf} is not unique. If $(\mathbf{U}^{*}, \mathbf{H}^{*})$ represents an optimal solution and there exist two matrices $\mathbf{R}_1$ and $\mathbf{R}_2$ such that $\mathbf{U}^{*} \mathbf{R}_1 \geq 0$ and $\mathbf{H}^{*} \mathbf{R}_2 \geq 0$, respectively, then the pair $(\mathbf{U}^{*}\mathbf{R}_1, \mathbf{H}^{*}\mathbf{R}_2)$ also achieves the same objective function value. The existence of multiple solutions further leads to different clustering results. 

To address this issue, we introduce the following orthogonal KCF (OKCF) \cite{yang2022ecca},
\begin{align}
	\min_{\mathbf{U}, \mathbf{H}} \quad & \tr(\mathbf{K}) - 2 \tr(\mathbf{U}^T\mathbf{K} \mathbf{H}) + \tr(\mathbf{U}^T\mathbf{K}\mathbf{U} )  \\
	\text{s.t.} \quad & \mathbf{U} \geq 0, \quad \mathbf{H}^T\mathbf{H} = \mathbf{I},\nonumber
\end{align}
where the non-negative constraint on $\mathbf{H}$ is replaced with the orthogonal constraint. Unlike traditional KCF, OKCF reduces the degree of freedom of the decomposition, enhancing clustering performance and interpretation. In OKCF, $\mathbf{H}$ is solely involved in the second term due to this constraint, making it more appealing for optimization.

To address the challenge of handling multiple kernels, we propose the Efficient Multiple Kernel Consensus Factorization method. This method integrates the consensus representation $\mathbf{U}$ and kernel-specific factor matrices $\{\mathbf{H}_r\}_{r=1}^{m}$ within a weighted sum of squared loss terms, as defined in Eq.~\eqref{eemkcf}.
\begin{align}\label{eemkcf}
	\min \quad & \sum_{r=1}^{m} \frac{1}{\mu_r}\left( \tr(\mathbf{K}_r) - 2 \tr(\mathbf{U}^T\mathbf{K}_r \mathbf{H}_r) + \tr(\mathbf{U}^T\mathbf{K}_r\mathbf{U} ) \right)  \nonumber \\
	\text{s.t.} \quad & \mathbf{U} \geq 0,  \mathbf{H}_r^T \mathbf{H}_r = \mathbf{I},  \sum_{r=1}^{m} \mu_r = 1,  \mu_r \geq 0,  \forall r,
\end{align}
where the weight $\frac{1}{\mu_r}$ reflects the importance for each kernel. 

The primary objective of EMKCF is to capture the shared underlying structure present in all kernels through the consensus representation $\mathbf{U}$. The optimization process aims to minimize the discrepancy between this consensus representation and the data across all kernels. Each kernel is associated with its own orthogonal factor matrix $\mathbf{H}_r$, which captures kernel-specific information. These factor matrices are constrained to be orthogonal, ensuring that they represent independent and complementary aspects of the data.

By jointly optimizing $\mathbf{U}$ and $\mathbf{H}_r$ across all kernels, EMKCF aims to find a consensus representation that effectively captures the underlying structure present in the multi-kernel data. This approach facilitates enhanced analysis and interpretation of complex datasets with multiple sources of information.
\subsection{Optimization}
As demonstrated in Eq. \eqref{eemkcf}, EEMKCF encompasses three type of variables $\{\mathbf{H}_r\}_{r=1}^{m}, \{\mu_r\}_{r=1}^{m}$, and $\mathbf{U}$. In this subsection we present an iterative optimization algorithm based on the block coordinate descent strategy. The specific process is outlined as follows.

\subsubsection{Update $\mathbf{H}_r$} The problem of Eq. \eqref{eemkcf} w.r.t. $\mathbf{H}_r$ becomes,
\begin{align}\label{opt_1}
	\max_{\mathbf{H}_r} \quad & \tr\left(\mathbf{H}_r^{T} \mathbf{E}_r\right), \quad \st \quad  \mathbf{H}_r^{T} \mathbf{H}_{r}=\mathbf{I}_c,
\end{align}
where $\mathbf{E}_{r} = \mathbf{K}_{r} \mathbf{U}$. By considering the singular value decomposition (SVD) on the matrix $\mathbf{E}_r$, i.e., $\mathbf{E}_r = \mathbf{P}_r \mathbf{Q}_r \mathbf{R}_r^{T}$, the optimal solution of $\mathbf{H}_r$ can be expressed as, 
\begin{align}\label{update_H}
	\mathbf{H}_r=\mathbf{P}_r \mathbf{R}_r^{T}. 
\end{align}
The specific proof process can be found in reference \cite{huang2014robust}.

\subsubsection{Update $\boldsymbol{\mu}$} The problem of Eq. \eqref{eemkcf} w.r.t. $\boldsymbol{\mu}$ becomes,
\begin{align}\label{opt_2}
	\min_{\boldsymbol{\mu}} \quad & \sum_{r=1}^m \frac{\beta_r}{\mu_r}  \quad \st \quad \sum_{r=1}^m \mu_r=1, \mu_r \geq 0,
\end{align}
where $\beta_r= \tr(\mathbf{K}_r) - 2 \tr(\mathbf{U}^T\mathbf{K}_r \mathbf{H}_r) + \tr(\mathbf{U}^T\mathbf{K}_r\mathbf{U} )  $. According to the Cauchy-Schwarz Inequality, the optimal solution of Eq. \eqref{opt_2} is achieved at 
\begin{equation}\label{update_mu}
	\mu_r=\frac{\sqrt{\beta_r}}{\sum_{r=1}^m \sqrt{\beta_r}}.
\end{equation}

\subsubsection{Update $\mathbf{U}$} The problem of Eq. \eqref{eemkcf} w.r.t. $\mathbf{U}$ becomes,
\begin{align}\label{opt_3}
	\min_{\mathbf{U}}  \quad & \tr(\mathbf{U}^T\mathbf{A} \mathbf{U} ) + 2\tr(\mathbf{B}^T\mathbf{U}) \\ 
	\st \quad  & \mathbf{U} \geq 0, \nonumber
\end{align}
where $\mathbf{A} = \sum_{r=1}^{m}\frac{1}{\mu_r} \mathbf{K}_r$ and $\mathbf{B} = -\sum_{r=1}^{m} \frac{1}{\mu_r} \mathbf{K}_r\mathbf{H}_r$.

Let $\mathbf{A} = \mathbf{A}^{+} - \mathbf{A}^{-}$ where $\mathbf{A}^{+}$ and $\mathbf{A}^{-}$ are two symmetric matrices whose elements are all positive. According to the theorem 1 in \cite{xu2004document}, the multiplicative update rule of $\mathbf{U}$ can be obtained as follows,
\begin{align}\label{update_U}
	\mathbf{U}_{ij} = \mathbf{U}_{ij}\left[ \frac{- \mathbf{B}_{ij} + \sqrt{\mathbf{B}_{ij}^2 + 4(\mathbf{A}^{+} \mathbf{U})_{ij}(\mathbf{A}^{-} \mathbf{U})_{ij}}}{2 (\mathbf{A}^{+} \mathbf{U})_{ij}} \right].
\end{align}

In summary, Algorithm \ref{eemkcf} outlines the entire procedure of the proposed method.

\begin{algorithm}[htbp]
	\caption{Algorithm for EMKCF in Eq.~\eqref{eemkcf}.}
	\label{alg-EEMKCF-1}
	\begin{algorithmic}[1] % Specify 1 to number every line
		\REQUIRE{Base kernel matrices $\{\mathbf{K}_r^o\}_{r=1}^{m}$, the cluster number $c$, the neighbor size $k=15$.}
		\STATE{Compute $\{\mathbf{S}_r\}_{r=1}^{m}$ according to Eq. \eqref{lnwkr};}
		\STATE{Compute $\{\mathbf{K}_r\}_{r=1}^{m}$ according to Eq. \eqref{new_kerenl};}		
		\STATE{Initialize $\boldsymbol{\mu},\mathbf{U}$;}
		\REPEAT
		\STATE{Update $\{\mathbf{H}_r\}_{r=1}^{m}$ according to Eq.~\eqref{update_H};}
		\STATE{Update $\boldsymbol{\mu}$ according to Eq.~\eqref{update_mu};}
		\STATE{Update $\mathbf{U}$ according to Eq.~\eqref{update_U};}
		\UNTIL{Converges}
		\ENSURE{Run clustering result from $\mathbf{U}$.}
	\end{algorithmic}
\end{algorithm}

\subsection{Convergence and Complexity Analysis}

The optimization problem in Eq.\eqref{eemkcf} is evidently lower-bounded, and the variable updates defined by Eqs.\eqref{update_H}, \eqref{update_mu}, and \eqref{update_U} monotonically decrease the objective function. Thus, convergence of Algorithm \ref{alg-EEMKCF-1} is guaranteed.

The time complexity for computing $\{\mathbf{K}_r\}_{r=1}^{m}$ according to Eq.~\eqref{new_kerenl} is $\mathcal{O}(mn^2k)$, primarily due to the sorting operation. Updating $\{\mathbf{K}_r\}_{r=1}^{m}$  requires $\mathcal{O}(mnkc + mnc^2)$ for matrix multiplication and SVD. Updating $\boldsymbol{\beta}$ requires $\mathcal{O}(mnc)$ for matrix multiplication. Updating $\mathbf{U}$ requires $\mathcal{O}(mnkc)$ for matrix multiplication. Thus, the total time complexity of the algorithm is $\mathcal{O}(mn^2k + (mnkc+mnc^2)t)$, where $t$ is the number of iterations. Notably, the complexity of EMKCF is linear w.r.t. $\mathcal{O}(n)$, discounting the sorting operation.

Since local kernel regression can be performed on each row of each kernel, the memory cost for computing $\{\mathbf{K}_r\}_{r=1}^{m}$  is only $\mathcal{O}(mnk)$. The variables during iterations require $\mathcal{O}(mnc)$ memory. Consequently, the total memory cost is also linear w.r.t. $\mathcal{O}(n)$. % Therefore, the proposed method is suitable for large-scale MKC tasks.

\section{Experiments}
In this section, extensive experiments are conducted to verify the effectiveness of the proposed method.
\subsection{Datasets}

We evaluate our method on seven benchmark datasets, spanning single-cell RNA-seq (Trachea, Fat, CITECBMC), general images datasets (Caltech101, 100Leaves), and large-scale datasets (MNISTLarge, EMNIST). The sample sizes range from 1013 to 280000, and the number of clusters varies from 7 to 100. This diversity makes these datasets suitable as benchmarks for MKC. The detailed information about these datasets is provided in Table \ref{tab:datasets}.

\begin{table}[htbp]
	\centering
	\caption{Summary of ten benchmark multiple kernel datasets.}
	\label{tab:datasets}
	\setlength{\tabcolsep}{3pt}
	\begin{tabular}{@{}c@{\hspace{15pt}}ccccc@{}}
		\toprule
		ID & Dataset & $\#$ Samples & $\#$ Features & $\#$ Classes & $\#$ Kernels \\ \midrule
		D1 & Trachea    & 1013   & 13741   & 7     & 12     \\ 
		D2 & Caltech101      & 1474   & 48    & 7    & 12     \\ 
		D3 & 100Leaves      & 1600   & 64    & 100    & 12     \\ 
		D4 & Fat        & 3618   & 15492   & 9     & 
		12     \\
		D5 & CITECBMC   & 8617   & 1703    & 15    & 12     \\ 
		D6 & MNISTLarge & 70000  & 784     & 10    & 12     \\ 
		D7 & EMNIST     & 280000 & 784     & 10    & 12     \\
		\bottomrule
	\end{tabular}
\end{table}

\begin{table}[htbp]
	\centering
	\caption{Parameter settings for Large Scale MKC}
	\resizebox{0.48\textwidth}{!}{
		%\resizebox{\linewidth}{!}{
			%\scalebox{1}{
				\setlength{\tabcolsep}{6pt}	
				\begin{tabular}{ccc}
					\toprule
					Methods & Range of Parameters & Grids \\
					\midrule
					MKKMMR &$\lambda \in 2^{[-15:1:15]}$& 31 \\
					OPLFMVC & Hyper-parameter free & 1 \\
					LSMKKM &$\tau \in[0.05:0.05:0.95]$& 19 \\
					SMKKM & Hyper-parameter free & 1 \\			
					DPMKKM & Hyper-parameter free & 1 \\
					EMKC &$\alpha \in 2^{[-15:2:15]},\beta \in 2^{[-15:2:15]}$& 256 \\
					MKCCSA &$\alpha \in 2^{[-15:2:15]}, \textrm{anchor} \in [c, \max(2c, 50), \max(4c, 100)]$& 48 \\
					LFPGR &$\lambda \in[2^{-2},2^{-1},1,2,4],\beta \in[2.^{-2},2.^{-1},1,2,4]$& 25 \\
					% OSLR &$\textrm{kscale} \in[0.1:0.2:1.5],\textrm{rscale} \in[1:1:5],\alpha = 0.01$& 40 \\
					FAMKKM &$\lambda_1 \in[0.01,0.1,1],\lambda_2 \in[0.01,0.1,1]$& 9 \\
					%CMKC &$r=\max(c,3\sqrt{n})$& 1 \\			
					FMKC &$k=15$& 1 \\						
					\bottomrule
				\end{tabular}
			}
			\label{emkcf_setting}
		\end{table}

		% \usepackage{booktabs}
		\begin{table}[htbp]
			\centering
			\caption{ACC/NMI/Time results of compared algorithms on all datasets}
			\label{res_aio}
			\resizebox{0.48\textwidth}{!}{
				\setlength{\tabcolsep}{6pt}
				\begin{tabular}{@{}cccccccc@{}}
					\toprule
					Methods & D1    & D2    & D3    & D4    & D5    & D6    & D7    \\ \midrule
					\multicolumn{8}{c}{ACC($\%$)} \\ \midrule
					MKKMMR  & 58.64 & 30.66 & 56.38 & 55.91 & 44.40  & N/A   & N/A   \\
					OPLFMVC & 58.74 & 29.58 & 67.19 & 51.74 & 41.98 & 54.09 & N/A   \\
					LSMKKM  & 63.5  & 30.46 & 63.88 & 53.28 & 51.89 & N/A   & N/A   \\
					SMKKM   & 54.01 & 31.55 & 61.19 & 51.77 & 43.18 & N/A   & N/A   \\
					DPMKKM  & 42.05 & 34.12 & 65.56 & 41.51 & 49.99 & N/A   & N/A   \\
					EMKC    & 66.24 & \underline{36.34} & 62.36 & 64.21 & 42.27 & N/A   & N/A   \\
					MKCCSA  & \underline{67.72} & 36.16 & 63.44 & \underline{64.68} & \underline{59.14} & \underline{58.65} & N/A   \\
					LFPGR   & 58.88 & 32.63 & \underline{67.25} & 55.40  & 45.04 & N/A   & N/A   \\
					FAMKKM  & 58.24 & 31.75 & 59.81 & 60.36 & 43.74 & N/A   & N/A   \\
					\midrule
					EMKCF  & \textbf{71.44} & \textbf{37.05} & \textbf{68.38} &\textbf{ 72.64} & \textbf{61.80}  & \textbf{76.84} &\textbf{49.03}  \\ 		\midrule
					\multicolumn{8}{c}{NMI($\%$)} \\
					\midrule
					MKKMMR  & 55.20  & 17.99 & 76.62 & 50.81 & 51.97 & N/A   & N/A   \\
					OPLFMVC & 58.67 & 17.88 & \underline{83.02} & 49.34 & 50.62 & 47.81 & N/A   \\
					LSMKKM  & 58.14 & 18.34 & 80.79 & 52.41 & 58.67 & N/A   & N/A   \\
					SMKKM   & 49.62 & 16.22 & 79.69 & 48.87 & 52.37 & N/A   & N/A   \\
					DPMKKM  & 44.56 & 19.29 & 82.03 & 40.03 & 56.41 & N/A   & N/A   \\
					EMKC    & \underline{61.01} & \underline{19.81}  & 74.68 & 60.38 & 49.06 & N/A   & N/A   \\
					MKCCSA  & 60.79 & 18.47 & 80.95 & \underline{62.20} & \underline{60.10}  & \underline{58.01} & N/A   \\
					LFPGR   & 58.29 & 19.36 & 82.67 & 52.03 & 51.90  & N/A   & N/A   \\
					FAMKKM  & 54.87 & 17.94 & 79.10 & 52.51 & 50.94 & N/A   & N/A   \\
					EMKCF  & \textbf{69.23} & \textbf{21.90} & \textbf{83.41} &\textbf{68.14}  &\textbf{66.42}  & \textbf{74.51} & \textbf{48.64} \\ \midrule
					\multicolumn{8}{c}{Time(s)} \\
					\midrule
					Methods & D1    & D2   & D3   & D4     & D5      & D6     & D7    \\ \midrule
					MKKMMR  & 62.1  &\textbf{0.8}   &\textbf{4.4}   & 1984.0 & 19021.3 & N/A    & N/A   \\
					OPLFMVC & 1.6   & 1.2  & 18.0 & 11.7   &\underline{38.9}   & \underline{1162.4} & N/A   \\
					LSMKKM  & 60.2  & 3.9  & 12.0 & 2353.1 & 30976.3 & N/A    & N/A   \\
					SMKKM   & 10.2  & 26.3 & 46.1 & 108.5  & 2558.6  & N/A    & N/A   \\
					DPMKKM  & \textbf{0.6}   & 1.8  & 33.7 & \underline{10.2}    & 92.4    & N/A    & N/A   \\
					EMKC    & 115.5 & 2.3  & 23.7 & 1348.3 & 6273.9  & N/A    & N/A   \\
					MKCCSA  & 39.0  & 13.4 & 15.6 & 555.6  & 2560.0  & 7522.8 & N/A   \\
					LFPGR   & 22.8  & 5.3  & 8.1  & 352.7  & 4797.2  & N/A    & N/A   \\
					FAMKKM  & 12.9  & 1.4  & \underline{4.6} & 226.1  & 2640.4  & N/A    & N/A   \\
					EMKCF  & \underline{0.7}   & \underline{1.0}  & 29.1 & \textbf{8.1}   &\textbf{16.2}  & \textbf{47.4}  & \textbf{212.0} \\ \bottomrule
				\end{tabular}
			}
		\end{table}

		\subsection{Compared Methods}
		We evaluated our method by comparing it with ten leading MKC clustering approaches widely recognized in MKC: MKKMMR \cite{mkkmmr_aaai_2016}, OPLFMVC \cite{liu2021one}, LSMKKM \cite{LSMKKM_2021}, SMKKM \cite{SMKKM_TPAMI_2023}, DPMKKM \cite{DPMKKM_tip_2022}, EMKC \cite{EMKC}, MKCCSA \cite{CSAMKC}, LFPGR \cite{lfpgr_tnnls_2023}, FAMKKM \cite{FAMKKM}. %These methods encompass kernel subspace clustering, kernel local graph approaches, late fusion techniques, and parameter-free methods, collectively providing a comprehensive evaluation that sheds light on the performance and its comparative position of BMKC in the field. 

		\subsection{Experimental Settings}
		
		For each dataset, we generated 12 kernels following established practices in MKC literature \cite{rmkkm_ijcai_2015}. We obtained the public source codes of all compared methods from their respective authors and fine-tuned the hyperparameters to optimize clustering performance, as detailed in Table \ref{emkcf_setting}. Moreover, our proposed method maintains a fixed nearest neighbor size of 15 without any tuning. We evaluate the performance using Accuracy (ACC), Normalized Mutual Information (NMI), and tuning time, with the results presented in Tables \ref{res_aio}. Given that all methods utilized Matlab for experimentation, we conducted our experiments on a PC equipped with an Intel Core-i7-10700 CPU and 128GB RAM, utilizing Matlab R2020a.

		\subsection{Clustering Results Analysis}
		
		In Tables \ref{res_aio}, the best clustering results on each dataset are bolded, and N/A denotes out-of-memory failure. According to the results, we can obtain the following observations.
		
		The clustering results presented in Table \ref{res_aio} offer valuable insights into the performance of various algorithms across different benchmark datasets. Notably, our proposed method consistently outperforms other algorithms in terms of accuracy (ACC) and normalized mutual information (NMI). For instance, on the Trachea dataset (D1), our method achieves an impressive ACC score of 71.44$\%$, surpassing all other methods. Similarly, on the MNISTLarge dataset (D6), our method achieves the highest ACC score of 76.84$\%$, indicating its robustness in handling large-scale datasets with complex clustering structures. Additionally, our method demonstrates remarkable scalability, achieving competitive results on datasets with varying sample sizes, features, and class distributions.
		
		Moreover, our method exhibits notable efficiency, as evidenced by its significantly lower time consumption compared to alternative algorithms. For instance, on the EMNIST dataset (D7), our method achieves superior clustering performance while requiring only 212.0 seconds for computation, in contrast to other methods that exhibit substantially longer processing times. This highlights the computational effectiveness of our approach, making it suitable for real-world applications where processing time is a critical factor.
		
		%Overall, the experimental results underscore the superiority of our proposed method in terms of both clustering accuracy and computational efficiency, positioning it as a promising solution for large-scale clustering tasks across diverse datasets.
		
		\subsection{Convergence Analysis}
\begin{figure}[htbp]
	\centering
	% Subfigure A
	\subfloat[Convergence on D1]{%
		\includegraphics[width=0.23\textwidth]{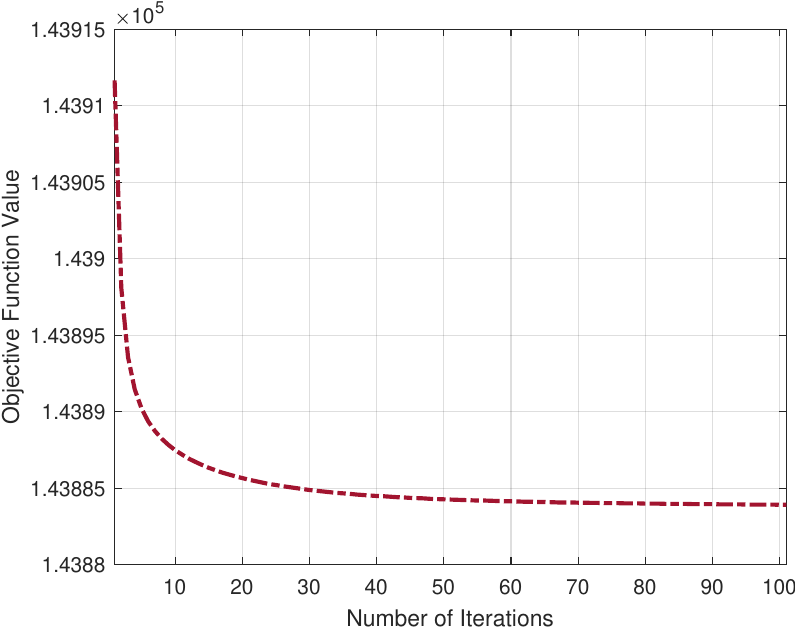}%
		\label{fig:D1}%
	}
	\hspace{0.1in}
	% Subfigure B
	\subfloat[Convergence on D4]{%
		\includegraphics[width=0.235\textwidth]{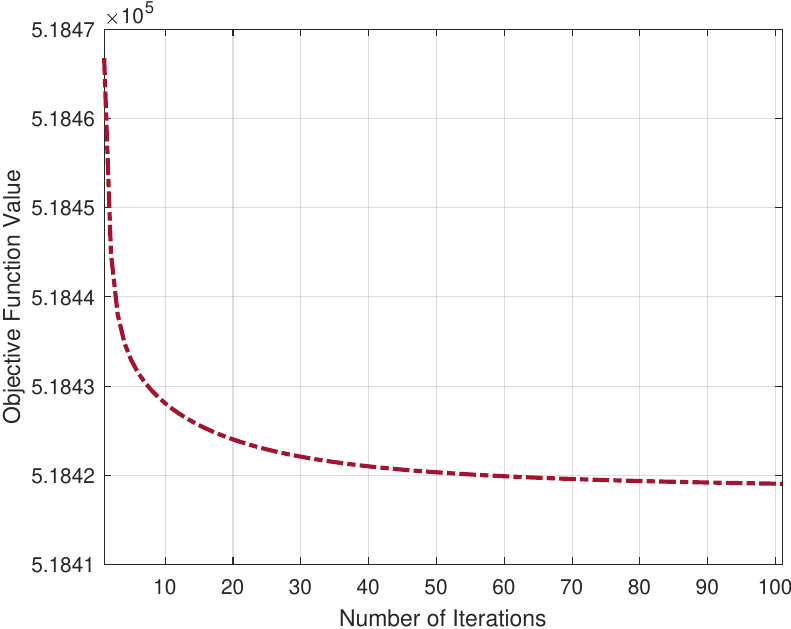}%
		\label{fig:D4}%
	}
	\caption{The convergence of EMKFC on D1 and D4.}
	\label{converges}
\end{figure}

		This section analyzes the convergence of EMKCF by tracking the variation of the objective function  with iterations. Figure \ref{converges} summarizes convergence trends on D1 and D4 datasets, revealing consistent monotonic decreases, indicating convergence.
		
		\section{Conclusion}
		MKC offers a promising solution to traditional kernel method challenges, particularly in nonlinear data clustering. Our investigation highlights limitations in both early and late fusion methods regarding memory usage and computational efficiency. To address this, we propose EMKCF, which optimizes computation and storage through sparse kernel matrix construction and extended orthogonal concept factorization. Experimental results across datasets demonstrate its superior scalability and effectiveness compared to SOTA methods.
%\section*{References}

\bibliographystyle{IEEEtran}
\bibliography{EEMKCF-2024.bib}

\begin{thebibliography}{10}
\providecommand{\url}[1]{#1}
\csname url@samestyle\endcsname
\providecommand{\newblock}{\relax}
\providecommand{\bibinfo}[2]{#2}
\providecommand{\BIBentrySTDinterwordspacing}{\spaceskip=0pt\relax}
\providecommand{\BIBentryALTinterwordstretchfactor}{4}
\providecommand{\BIBentryALTinterwordspacing}{\spaceskip=\fontdimen2\font plus
\BIBentryALTinterwordstretchfactor\fontdimen3\font minus
  \fontdimen4\font\relax}
\providecommand{\BIBforeignlanguage}[2]{{%
\expandafter\ifx\csname l@#1\endcsname\relax
\typeout{** WARNING: IEEEtran.bst: No hyphenation pattern has been}%
\typeout{** loaded for the language `#1'. Using the pattern for}%
\typeout{** the default language instead.}%
\else
\language=\csname l@#1\endcsname
\fi
#2}}
\providecommand{\BIBdecl}{\relax}
\BIBdecl

\bibitem{lmkkm_nips_2014}
M.~Gönen and A.~A. Margolin, ``Localized data fusion for kernel k-means
  clustering with application to cancer biology,'' \emph{NeurIPS}, pp.
  1305--1313, 2014.

\bibitem{wang2017visualization}
B.~Wang, J.~Zhu \emph{et~al.}, ``Visualization and analysis of single-cell
  rna-seq data by kernel-based similarity learning,'' \emph{Nature methods},
  vol.~14, no.~4, pp. 414--416, 2017.

\bibitem{lcsg_tii_2021}
Z.~Ren, M.~Mukherjee, J.~Lloret, and P.~Venu, ``Multiple kernel driven
  clustering with locally consistent and selfish graph in industrial iot,''
  \emph{IEEE Transactions on Industrial Informatics}, vol.~17, no.~4, pp.
  2956--2963, 2021.

\bibitem{mkc_jsac_2021}
Z.~Ren, M.~Mukherjee, M.~Bennis, and J.~Lloret, ``Multikernel clustering via
  non-negative matrix factorization tailored graph tensor over distributed
  networks,'' \emph{IEEE Journal on Selected Areas in Communications}, vol.~39,
  no.~7, pp. 1946--1956, 2021.

\bibitem{scmk_aaai_2017}
Z.~Kang, C.~Peng, and Q.~Cheng, ``Twin learning for similarity and clustering:
  {A} unified kernel approach,'' in \emph{AAAI}, 2017, pp. 2080--2086.

\bibitem{scmk_aaai_2018}
Z.~Kang, C.~Peng, Q.~Cheng, and Z.~Xu, ``Unified spectral clustering with
  optimal graph,'' in \emph{AAAI}, 2018, pp. 3366--3373.

\bibitem{smkl_ijcai_2018}
Z.~Kang, X.~Lu, J.~Yi, and Z.~Xu, ``Self-weighted multiple kernel learning for
  graph-based clustering and semi-supervised classification,'' in \emph{IJCAI},
  2018, pp. 2312--2318.

\bibitem{lkgr_kbs_2019}
K.~Zhao, L.~Wen, W.~Chen, and Z.~Xu, ``Low-rank kernel learning for graph-based
  clustering,'' \emph{Knowledge-Based Systems}, vol. 163, pp. 510--517, 2019.

\bibitem{jmksc_is_2019}
C.~Yang, Z.~Ren, Q.~Sun, M.~Wu, M.~Yin, and Y.~Sun, ``Joint correntropy metric
  weighting and block diagonal regularizer for robust multiple kernel subspace
  clustering,'' \emph{Information Sciences}, vol. 500, pp. 48--66, 2019.

\bibitem{ssmkc_if_2020}
S.~Zhou, E.~Zhu, X.~Liu, T.~Zheng, Q.~Liu, J.~Xia, and J.~Yin, ``Subspace
  segmentation-based robust multiple kernel clustering,'' \emph{Information
  Fusion}, vol.~53, pp. 145--154, 2020.

\bibitem{cagl_tcyb_2021}
Z.~Ren, S.~X. Yang, Q.~Sun, and T.~Wang, ``Consensus affinity graph learning
  for multiple kernel clustering,'' \emph{IEEE Transactions on Cybernetics},
  vol.~51, no.~6, pp. 3273--3284, 2021.

\bibitem{lswmkc_tnnls_2022}
L.~Li, S.~Wang, X.~Liu, E.~Zhu, L.~Shen, K.~Li, and K.~Li, ``Local
  sample-weighted multiple kernel clustering with consensus discriminative
  graph,'' \emph{IEEE TNNLS}, pp. 1--14, 2022.

\bibitem{mkcss_tnnls_2020}
S.~Zhou, X.~Liu, M.~Li, E.~Zhu, L.~Liu, C.~Zhang, and J.~Yin, ``Multiple kernel
  clustering with neighbor-kernel subspace segmentation,'' \emph{IEEE TNNLS},
  vol.~31, no.~4, pp. 1351--1362, 2020.

\bibitem{lfpgr_tnnls_2023}
S.~Wang, X.~Liu, L.~Liu, S.~Zhou, and E.~Zhu, ``Late fusion multiple kernel
  clustering with proxy graph refinement,'' \emph{IEEE TNNLS}, vol.~34, no.~8,
  pp. 4359--4370, 2023.

\bibitem{SNC_2017}
X.~Chen, F.~Nie, J.~Z. Huang, and M.~Yang, ``Scalable normalized cut with
  improved spectral rotation.'' in \emph{IJCAI}, 2017, pp. 1518--1524.

\bibitem{DPMKKM_tip_2022}
R.~Wang, J.~Lu, Y.~Lu, F.~Nie, and X.~Li, ``Discrete and parameter-free
  multiple kernel k-means,'' \emph{IEEE Transactions on Image Processing},
  vol.~31, pp. 2796--2808, 2022.

\bibitem{Expcut_2021}
D.~Wu, F.~Nie, J.~Lu, R.~Wang, and X.~Li, ``Balanced graph cut with exponential
  inter-cluster compactness,'' \emph{IEEE Transactions on Artificial
  Intelligence}, vol.~3, no.~4, pp. 498 -- 505, 2022.

\bibitem{LABIN_2020}
X.~Chen, R.~Chen, Q.~Wu, Y.~Fang, F.~Nie, and J.~Z. Huang, ``Labin: Balanced
  min cut for large-scale data,'' \emph{IEEE TNNLS}, vol.~31, no.~3, pp.
  725--736, 2020.

\bibitem{LKM_2018}
Z.~Li, F.~Nie, X.~Chang, Z.~Ma, and Y.~Yang, ``Balanced clustering via
  exclusive lasso: A pragmatic approach,'' in \emph{AAAI}, vol.~32, 2018, pp.
  3596--3603.

\bibitem{BKNC_2022}
H.~Chen, Q.~Zhang, R.~Wang, F.~Nie, and X.~Li, ``A general soft-balanced
  clustering framework based on a novel balance regularizer,'' \emph{Signal
  Processing}, vol. 198, p. 108572, 2022.

\bibitem{EBMC_2020}
X.~Chen, W.~Hong, F.~Nie, J.~Z. Huang, and L.~Shen, ``Enhanced balanced min
  cut,'' \emph{International Journal of Computer Vision}, vol. 128, pp.
  1982--1995, 2020.

\bibitem{LGBC_2019}
J.~Han, H.~Liu, and F.~Nie, ``A local and global discriminative framework and
  optimization for balanced clustering,'' \emph{IEEE TNNLS}, vol.~30, no.~10,
  pp. 3059--3071, 2019.

\bibitem{rmkkm_ijcai_2015}
L.~Du, P.~Zhou \emph{et~al.}, ``Robust multiple kernel k-means using
  l21-norm,'' in \emph{IJCAI}, 2015, pp. 3476--3482.

\bibitem{mkkmmr_aaai_2016}
X.~Liu, Y.~Dou, J.~Yin, L.~Wang, and E.~Zhu, ``Multiple kernel \emph{k}-means
  clustering with matrix-induced regularization,'' in \emph{AAAI}, 2016, pp.
  1888--1894.

\bibitem{onkc_aaai_2017}
X.~Liu, S.~Zhou, Y.~Wang, M.~Li, Y.~Dou, E.~Zhu, and J.~Yin, ``Optimal
  neighborhood kernel clustering with multiple kernels,'' in \emph{AAAI}, 2017,
  pp. 2266--2272.

\bibitem{smkkm_tpami_2023}
X.~Liu, ``Simplemkkm: Simple multiple kernel k-means,'' \emph{IEEE TPAMI},
  vol.~45, no.~4, pp. 5174 -- 5186, 2023.

\bibitem{hmkc_aaai_2021}
J.~{Liu}, X.~{Liu}, S.~{Wang}, S.~{Zhou}, and Y.~{Yang}, ``Hierarchical
  multiple kernel clustering.'' in \emph{AAAI}, 2021, pp. 8671--8679.

\bibitem{lkamkc_ijcai_2016}
M.~Li, X.~Liu, L.~Wang, Y.~Dou, J.~Yin, and E.~Zhu, ``Multiple kernel
  clustering with local kernel alignment maximization,'' in \emph{IJCAI}, 2016,
  pp. 1704--1710.

\bibitem{onalk_tkde_2022}
J.~Liu, X.~Liu, J.~Xiong, Q.~Liao, S.~Zhou, S.~Wang, and Y.~Yang, ``Optimal
  neighborhood multiple kernel clustering with adaptive local kernels,''
  \emph{IEEE TKDE}, vol.~34, no.~6, pp. 2872--2885, 2022.

\bibitem{liu2023hyperparameter}
X.~Liu, ``Hyperparameter-free localized simple multiple kernel k-means with
  global optimum,'' \emph{IEEE TPAMI}, 2023.

\bibitem{llmkl_kbs_2020}
Z.~Ren, H.~Li, C.~Yang, and Q.~Sun, ``Multiple kernel subspace clustering with
  local structural graph and low-rank consensus kernel learning,''
  \emph{Knowledge-Based Systems}, vol. 188, p. 105040, 2020.

\bibitem{slmkc_is_2021}
Z.~Ren, H.~Lei, Q.~Sun, and C.~Yang, ``Simultaneous learning coefficient matrix
  and affinity graph for multiple kernel clustering,'' \emph{Information
  Sciences}, vol. 547, pp. 289--306, 2021.

\bibitem{sgmk_pr_2021}
Z.~Kang, C.~Peng, Q.~Cheng, X.~Liu, X.~Peng, Z.~Xu, and L.~Tian, ``Structured
  graph learning for clustering and semi-supervised classification,''
  \emph{Pattern Recognition}, vol. 110, p. 107627, 2021.

\bibitem{spmkc_tnnls_2021}
Z.~Ren and Q.~Sun, ``Simultaneous global and local graph structure preserving
  for multiple kernel clustering,'' \emph{IEEE TNNLS}, vol.~32, no.~5, pp.
  1839--1851, 2021.

\bibitem{pgmkc_neuro_2021}
X.~Li, Z.~Ren, H.~Lei, Y.~Huang, and Q.~Sun, ``Multiple kernel clustering with
  pure graph learning scheme,'' \emph{Neurocomputing}, vol. 424, pp. 215--225,
  2021.

\bibitem{kcgt_aaai_2021}
Z.~Ren, Q.~Sun, and D.~Wei, ``Multiple kernel clustering with kernel k-means
  coupled graph tensor learning,'' in \emph{AAAI}, vol.~35, no.~11, 2021, pp.
  9411--9418.

\bibitem{you2023clustering}
J.~You, C.~Han, Z.~Ren, H.~Li, and X.~You, ``Clustering via multiple kernel
  k-means coupled graph and enhanced tensor learning,'' \emph{Applied
  Intelligence}, vol.~53, no.~3, pp. 2564--2575, 2023.

\bibitem{wang2019multi}
S.~{Wang}, X.~{Liu}, E.~{Zhu}, C.~{Tang}, J.~{Liu}, J.~{Hu}, J.~{Xia}, and
  J.~{Yin}, ``Multi-view clustering via late fusion alignment maximization,''
  in \emph{IJCAI}, 2019, pp. 3778--3784.

\bibitem{liu2021one}
X.~Liu, L.~Liu, Q.~Liao, S.~Wang, Y.~Zhang, W.~Tu, C.~Tang, J.~Liu, and E.~Zhu,
  ``One pass late fusion multi-view clustering,'' in \emph{ICML}, vol. 139,
  2021, pp. 6850--6859.

\bibitem{csamkc_tnnls_2023}
S.~Zhou, Q.~Ou, X.~Liu, S.~Wang, L.~Liu, S.~Wang, E.~Zhu, J.~Yin, and X.~Xu,
  ``Multiple kernel clustering with compressed subspace alignment,'' \emph{IEEE
  TNNLS}, vol.~34, no.~1, pp. 252--263, 2023.

\bibitem{c3lmc_tcsvt_2023}
X.~Li, Y.~Sun, Q.~Sun, and Z.~Ren, ``Consensus cluster center guided latent
  multi-kernel clustering,'' \emph{IEEE Transactions on Circuits and Systems
  for Video Technology}, vol.~33, no.~6, pp. 2864--2876, 2023.

\bibitem{lflka_tmm_2023}
T.~Zhang, X.~Liu, L.~Gong, S.~Wang, X.~Niu, and L.~Shen, ``Late fusion multiple
  kernel clustering with local kernel alignment maximization,'' \emph{IEEE
  Transactions on Multimedia}, vol.~25, pp. 993--1007, 2023.

\bibitem{zhang2022fusion}
Y.~Zhang, X.~Liu, J.~Liu, S.~Dai, C.~Zhang, K.~Xu, and E.~Zhu, ``Fusion
  multiple kernel k-means,'' in \emph{AAAI}, vol.~36, no.~8, 2022, pp.
  9109--9117.

\bibitem{pmksc_tmm_2022}
M.~Sun, S.~Wang, P.~Zhang, X.~Liu, X.~Guo, S.~Zhou, and E.~Zhu, ``Projective
  multiple kernel subspace clustering,'' \emph{IEEE Transactions on
  Multimedia}, vol.~24, pp. 2567--2579, 2022.

\bibitem{ckm_2000}
P.~S. Bradley, K.~P. Bennett, and A.~Demiriz, ``Constrained k-means
  clustering,'' \emph{Microsoft Research, Redmond}, vol.~20, no.~0, p.~0, 2000.

\bibitem{bkm_2014}
M.~I. Malinen and P.~Fr{\"a}nti, ``Balanced k-means for clustering,'' in
  \emph{Structural, Syntactic, and Statistical Pattern Recognition: Joint IAPR
  International Workshop}, 2014, pp. 32--41.

\bibitem{bkm_2018}
H.~M. Le, A.~Eriksson, T.-T. Do, and M.~Milford, ``A binary optimization
  approach for constrained k-means clustering,'' in \emph{ACCV}, 2018, pp.
  383--398.

\bibitem{bkm_2019}
W.~Tang, Y.~Yang, L.~Zeng, and Y.~Zhan, ``Optimizing mse for clustering with
  balanced size constraints,'' \emph{Symmetry}, vol.~11, no.~3, p. 338, 2019.

\bibitem{vns_2017}
L.~R. Costa, D.~Aloise, and N.~Mladenovi{\'c}, ``Less is more: basic variable
  neighborhood search heuristic for balanced minimum sum-of-squares
  clustering,'' \emph{Information Sciences}, vol. 415, pp. 247--253, 2017.

\bibitem{so_2022}
R.~Martin-Santamaria, J.~S{\'a}nchez-Oro, S.~P{\'e}rez-Pel{\'o}, and A.~Duarte,
  ``Strategic oscillation for the balanced minimum sum-of-squares clustering
  problem,'' \emph{Information Sciences}, vol. 585, pp. 529--542, 2022.

\bibitem{soft_km_2002}
A.~Banerjee and J.~Ghosh, ``On scaling up balanced clustering algorithms,'' in
  \emph{SIAM ICDM}, 2002, pp. 333--349.

\bibitem{zhou2010exclusive}
Y.~Zhou, R.~Jin, and S.~C.-H. Hoi, ``Exclusive lasso for multi-task feature
  selection,'' in \emph{IJCAI}, 2010, pp. 988--995.

\bibitem{BCLS_2017}
H.~Liu, J.~Han, F.~Nie, and X.~Li, ``Balanced clustering with least square
  regression,'' in \emph{AAAI}, vol.~31, no.~1, 2017, pp. 2231--2237.

\bibitem{gbc_tnnls_2019}
J.~Han, H.~Liu, and F.~Nie, ``A local and global discriminative framework and
  optimization for balanced clustering,'' \emph{IEEE TNNLS}, vol.~30, no.~10,
  pp. 3059--3071, 2019.

\bibitem{FCFC_2018}
H.~Liu, Z.~Huang, Q.~Chen, M.~Li, Y.~Fu, and L.~Zhang, ``Fast clustering with
  flexible balance constraints,'' in \emph{IEEE International Conference on Big
  Data (Big Data)}, 2018, pp. 743--750.

\bibitem{RKM_2019}
W.~Lin, Z.~He, and M.~Xiao, ``Balanced clustering: A uniform model and fast
  algorithm.'' in \emph{IJCAI}, 2019, pp. 2987--2993.

\bibitem{cauchy_b_nmf_air_2023}
H.~Xiong, D.~Kong, and F.~Nie, ``Cauchy balanced nonnegative matrix
  factorization,'' \emph{Artificial Intelligence Review}, pp. 1--37, 2023.

\bibitem{submodular_prl_2011}
Y.~Kawahara, K.~Nagano, and Y.~Okamoto, ``Submodular fractional programming for
  balanced clustering,'' \emph{Pattern Recognition Letters}, vol.~32, no.~2,
  pp. 235--243, 2011.

\bibitem{bkcut_nips_2014}
S.~S. Rangapuram, P.~K. Mudrakarta, and M.~Hein, ``Tight continuous relaxation
  of the balanced k-cut problem,'' \emph{NeurIPS}, vol.~27, 2014.

\bibitem{ratiocut_1991}
L.~Hagen and A.~Kahng, ``Fast spectral methods for ratio cut partitioning and
  clustering,'' in \emph{IEEE international conference on computer-aided design
  digest}, 1991, pp. 10--11.

\bibitem{ncut_tpami_2000}
J.~Shi and J.~Malik, ``Normalized cuts and image segmentation,'' \emph{IEEE
  TPAMI}, vol.~22, no.~8, pp. 888--905, 2000.

\bibitem{mmcut_icdm_2001}
C.~H. Ding, X.~He, H.~Zha, M.~Gu, and H.~D. Simon, ``A min-max cut algorithm
  for graph partitioning and data clustering,'' in \emph{IEEE ICDM}, 2001, pp.
  107--114.

\bibitem{mtvc_nips_2013}
X.~Bresson, T.~Laurent, D.~Uminsky, and J.~Von~Brecht, ``Multiclass total
  variation clustering,'' \emph{NeurIPS}, vol.~26, pp. 1421--–1429, 2013.

\bibitem{SBMC_2017}
X.~Chen, J.~Zhexue~Haung, F.~Nie, R.~Chen, and Q.~Wu, ``A self-balanced min-cut
  algorithm for image clustering,'' in \emph{ICCV}, 2017, pp. 2061--2069.

\bibitem{FFCAG_2022}
F.~Nie, C.~Liu, R.~Wang, Z.~Wang, and X.~Li, ``Fast fuzzy clustering based on
  anchor graph,'' \emph{IEEE TFS}, vol.~30, no.~7, pp. 2375--2387, 2022.

\bibitem{GBFC_2023}
C.~Liu, F.~Nie, R.~Wang, and X.~Li, ``Graph based soft-balanced fuzzy
  clustering,'' \emph{IEEE TFS}, vol.~31, no.~6, pp. 2044--2055, 2023.

\bibitem{lle}
S.~T. Roweis and L.~K. Saul, ``Nonlinear dimensionality reduction by locally
  linear embedding,'' \emph{Science}, vol. 290, no. 5500, pp. 2323--2326, 2000.

\bibitem{saul2003think}
L.~K. Saul and S.~T. Roweis, ``Think globally, fit locally: unsupervised
  learning of low dimensional manifolds,'' \emph{JMLR}, vol.~4, no. Jun, pp.
  119--155, 2003.

\bibitem{gasteiger2019diffusion}
J.~Gasteiger, S.~Wei{\ss}enberger, and S.~G{\"u}nnemann, ``Diffusion improves
  graph learning,'' \emph{NeurIPS}, vol.~32, 2019.

\bibitem{zhou2020unsupervised}
P.~Zhou, J.~Chen, M.~Fan, L.~Du, Y.-D. Shen, and X.~Li, ``Unsupervised feature
  selection for balanced clustering,'' \emph{Knowledge-Based Systems}, vol.
  193, p. 105417, 2020.

\bibitem{zhou2022balanced}
P.~Zhou, J.~Chen, L.~Du, and X.~Li, ``Balanced spectral feature selection,''
  \emph{IEEE Transactions on Cybernetics}, vol.~53, no.~7, pp. 4232--4244,
  2023.

\bibitem{gdc}
J.~Gasteiger, S.~Wei\ss~enberger, and S.~G\"{u}nnemann, ``Diffusion improves
  graph learning,'' in \emph{NeurIPS}, vol.~32, 2019.

\bibitem{CDKM_2022}
F.~Nie, J.~Xue, D.~Wu, R.~Wang, H.~Li, and X.~Li, ``Coordinate descent method
  for $k$-means,'' \emph{IEEE TPAMI}, vol.~44, no.~5, pp. 2371--2385, 2022.

\bibitem{LSMKKM_2021}
X.~Liu, S.~Zhou, L.~Liu, C.~Tang, S.~Wang, J.~Liu, and Y.~Zhang, ``Localized
  simple multiple kernel k-means,'' in \emph{ICCV}, 2021, pp. 9293--9301.

\bibitem{zhou2019evolutionary}
Z.-H. Zhou, Y.~Yu, and C.~Qian, \emph{Evolutionary learning: Advances in
  theories and algorithms}.\hskip 1em plus 0.5em minus 0.4em\relax Springer,
  2019.

\end{thebibliography}


\begin{thebibliography}{10}
\providecommand{\url}[1]{#1}
\csname url@samestyle\endcsname
\providecommand{\newblock}{\relax}
\providecommand{\bibinfo}[2]{#2}
\providecommand{\BIBentrySTDinterwordspacing}{\spaceskip=0pt\relax}
\providecommand{\BIBentryALTinterwordstretchfactor}{4}
\providecommand{\BIBentryALTinterwordspacing}{\spaceskip=\fontdimen2\font plus
\BIBentryALTinterwordstretchfactor\fontdimen3\font minus
  \fontdimen4\font\relax}
\providecommand{\BIBforeignlanguage}[2]{{%
\expandafter\ifx\csname l@#1\endcsname\relax
\typeout{** WARNING: IEEEtran.bst: No hyphenation pattern has been}%
\typeout{** loaded for the language `#1'. Using the pattern for}%
\typeout{** the default language instead.}%
\else
\language=\csname l@#1\endcsname
\fi
#2}}
\providecommand{\BIBdecl}{\relax}
\BIBdecl

\bibitem{mkkmmr}
X.~{Liu}, Y.~{Dou}, J.~{Yin}, L.~{Wang}, and E.~{Zhu}, ``Multiple kernel k
  -means clustering with matrix-induced regularization,'' in \emph{AAAI}, 2016,
  pp. 1888--1894.

\bibitem{SMKKM_TPAMI_2023}
X.~Liu, ``Simplemkkm: Simple multiple kernel k-means,'' \emph{IEEE Transactions
  on Pattern Analysis and Machine Intelligence}, vol.~45, no.~4, pp.
  5174--5186, 2023.

\bibitem{LSMKKM_2021}
X.~Liu, S.~Zhou, L.~Liu, C.~Tang, S.~Wang, J.~Liu, and Y.~Zhang, ``Localized
  simple multiple kernel k-means,'' in \emph{Proceedings of the IEEE/CVF
  International Conference on Computer Vision}, 2021, pp. 9293--9301.

\bibitem{DPMKKM_tip_2022}
R.~Wang, J.~Lu, Y.~Lu, F.~Nie, and X.~Li, ``Discrete and parameter-free
  multiple kernel k-means,'' \emph{IEEE Transactions on Image Processing},
  vol.~31, pp. 2796--2808, 2022.

\bibitem{EMKC}
C.~Tang, Z.~Li, W.~Yan, G.~Yue, and W.~Zhang, ``Efficient multiple kernel
  clustering via spectral perturbation,'' in \emph{Proceedings of the 30th ACM
  International Conference on Multimedia}, 2022, p. 1603–1611.

\bibitem{FAMKKM}
J.~Wang, C.~Tang, X.~Zheng, X.~Liu, W.~Zhang, E.~Zhu, and X.~Zhu, ``Fast
  approximated multiple kernel k-means,'' \emph{IEEE Transactions on Knowledge
  and Data Engineering}, no.~01, pp. 1--10, dec 5555.

\bibitem{liu2021one}
X.~Liu, L.~Liu, Q.~Liao, S.~Wang, Y.~Zhang, W.~Tu, C.~Tang, J.~Liu, and E.~Zhu,
  ``One pass late fusion multi-view clustering,'' in \emph{International
  conference on machine learning}, vol. 139, 2021, pp. 6850--6859.

\bibitem{CSAMKC}
S.~Zhou, Q.~Ou, X.~Liu, S.~Wang, L.~Liu, S.~Wang, E.~Zhu, J.~Yin, and X.~Xu,
  ``Multiple kernel clustering with compressed subspace alignment,'' \emph{IEEE
  Transactions on Neural Networks and Learning Systems}, vol.~34, no.~1, pp.
  252--263, 2023.

\bibitem{lfpgr_tnnls_2023}
S.~Wang, X.~Liu, L.~Liu, S.~Zhou, and E.~Zhu, ``Late fusion multiple kernel
  clustering with proxy graph refinement,'' \emph{IEEE TNNLS}, vol.~34, no.~8,
  pp. 4359--4370, 2023.

\bibitem{benedetti1977nonparametric}
J.~K. Benedetti, ``On the nonparametric estimation of regression functions,''
  \emph{Journal of the Royal Statistical Society: Series B (Methodological)},
  vol.~39, no.~2, pp. 248--253, 1977.

\bibitem{xu2004document}
W.~Xu and Y.~Gong, ``Document clustering by concept factorization,'' in
  \emph{Proceedings of the 27th annual international ACM SIGIR conference on
  Research and development in information retrieval}, 2004, pp. 202--209.

\bibitem{yang2022ecca}
B.~Yang, X.~Zhang, F.~Nie, B.~Chen, F.~Wang, Z.~Nan, and N.~Zheng, ``Ecca:
  Efficient correntropy-based clustering algorithm with orthogonal concept
  factorization,'' \emph{IEEE Transactions on Neural Networks and Learning
  Systems}, vol.~34, no.~10, pp. 7377--7390, 2022.

\bibitem{huang2014robust}
J.~Huang, F.~Nie, H.~Huang, and C.~Ding, ``Robust manifold nonnegative matrix
  factorization,'' \emph{ACM Transactions on Knowledge Discovery from Data
  (TKDD)}, vol.~8, no.~3, pp. 1--21, 2014.

\bibitem{mkkmmr_aaai_2016}
X.~Liu, Y.~Dou, J.~Yin, L.~Wang, and E.~Zhu, ``Multiple kernel \emph{k}-means
  clustering with matrix-induced regularization,'' in \emph{Proceedings of the
  AAAI conference on artificial intelligence}, 2016, pp. 1888--1894.

\bibitem{rmkkm_ijcai_2015}
L.~Du, P.~Zhou \emph{et~al.}, ``Robust multiple kernel k-means using
  l21-norm,'' in \emph{Twenty-fourth international joint conference on
  artificial intelligence}, 2015, pp. 3476--3482.

\end{thebibliography}

\end{document}